%% file: main.tex
\documentclass[]{spie}  %>>> use for US letter paper
\usepackage{amsmath,amsfonts,amssymb}
\usepackage{graphicx}
\usepackage{tabularx}
\usepackage{csquotes}
\usepackage{siunitx}
\usepackage{caption}
\usepackage{subcaption}
\usepackage{booktabs,multirow}
\usepackage[acronym]{glossaries}
\usepackage[T1]{fontenc}
\glsdisablehyper

\usepackage[colorlinks=true, allcolors=blue]{hyperref}
\usepackage[noabbrev, capitalize]{cleveref}

\newcolumntype{K}{S[table-format=2.2]}
\newacronym{mlp}{MLP}{Multi-Layer Perceptron}

\title{Multi-Modal Fusion Transformer for Visual Question Answering in Remote Sensing}

\author{Tim Siebert$^*$}
\author{Kai Norman Clasen$^*$}
\author{Mahdyar Ravanbakhsh}
\author{Begüm Demir}
\affil{Technische Universitat Berlin, Einsteinufer 17, 10587, Berlin, Germany}

\authorinfo{Further author information: (Send correspondence to Tim Siebert)\\Tim Siebert: E-mail: siebert@tu-berlin.de\\
*Equal Contribution}

\begin{document} 
\maketitle
%  \href{https://wandb.ai/timbo554/vgars-code_RSVQAxBEN_Dataset/reports/Multi-Modal-Fusion-Transformer-in-RS-VQA--VmlldzoyMzUzOTUy?accessToken=19wj0c8ptf78uax3hsb0uxy50uwig1km0kj68jto4yzic8cg1zrbdo33gzrrsfkn}{\textbf{model training report}}

% For this submission we could shorten the introduction due to space
% Make sure that the terms used in abstract
% Approach/system are different to model/architecture
% Same for module; Do not use step but module
% dataset names are not required for such a submission but here we include it because we also extended one dataset
% Make sure to motivate the reasoning/effect for adding the other channels -> Better VQA performance for complex LCLU classes
% If we use the acronym make sure to use it in abstract, methodology, experiments and results
\begin{abstract}
With the new generation of satellite technologies, the archives of remote sensing (RS) images are growing very fast.
% If the introduction should be shorter, we
% could fuse the first sentence with the following sentence
To make the intrinsic information of each RS image easily accessible, visual question answering (VQA) has been introduced in RS. VQA allows a user to formulate a free-form question concerning the content of RS images to extract generic information.
It has been shown that the fusion of the input modalities (i.e., image and text) is crucial for the performance of VQA systems. 
Most of the current fusion approaches use modality-specific representations in their fusion modules instead of joint representation learning. 
% change fully utilize both input modalities to make it clear that we are focusing on the alignment
% To fully discover the underlying relation between the two modalities, learn the relation ...
% which is not possible with a static non-learnable combinoatin
However, to discover the underlying relation between both the image and question modality, the model is required to learn the joint representation instead of simply combining (e.g., concatenating, adding, or multiplying) the modality-specific representations. We propose a multi-modal transformer-based architecture to overcome this issue.
Our proposed architecture consists of three main modules: i) the feature extraction module for extracting the modality-specific features; ii) the fusion module, which leverages a user-defined number of multi-modal transformer layers of the VisualBERT model (VB); and iii) the classification module to obtain the answer. In contrast to recently proposed transformer-based models in RS VQA, the presented architecture (called \emph{VBFusion}) is not limited to specific questions, e.g., questions concerning pre-defined objects.
Experimental results obtained on the RSVQAxBEN and RSVQA-LR datasets (which are made up of RGB bands of Sentinel-2 images) demonstrate the effectiveness of VBFusion for VQA tasks in RS. To analyze the importance of using other spectral bands for the description of the complex content of RS images in the framework of VQA, we extend the RSVQAxBEN dataset to include all the spectral bands of Sentinel-2 images with 10m and 20m spatial resolution. Experimental results show the importance of utilizing these bands to characterize the land-use land-cover classes present in the images in the framework of VQA. The code of the proposed method is publicly available at \href{https://git.tu-berlin.de/rsim/multi-modal-fusion-transformer-for-vqa-in-rs}{https://git.tu-berlin.de/rsim/multi-modal-fusion-transformer-for-vqa-in-rs}.
\end{abstract}

% \todo[inline]{Simple combination of modalities rather than statically or non-learnable. After first use of "simple combination" add (concatentation, addition, multiplication in parentheses)}

% Include a list of keywords after the abstract 

\section{INTRODUCTION}
\label{sec:intro}  % \label{} allows reference to this section
\input{introduction}

\section{Proposed multi-modal transformer-based VQA architecture}
\label{sec:method}
\input{method}

\section{Dataset Description and Experimental Setup}
\label{sec:exp}
\input{experiments}

\section{CONCLUSION}
\label{sec:con}
\input{conclusion}

\section*{ACKNOWLEDGEMENT}

This work is funded by the European Research Council (ERC) through the ERC-2017-STG BigEarth Project under Grant 759764 and by the German Ministry for Economic Affairs and Climate Action through the AI-Cube Project under Grant 50EE2012B.

% References
\bibliography{report} % bibliography data in report.bib
\bibliographystyle{spiebib} % makes bibtex use spiebib.bst

% \section{Original Abstract}
% With the new generation of satellite technologies, the archives of remote sensing (RS) images are growing very fast. These images contain a vast amount of valuable information. To make the information easily accessible from remote sensing images, one can utilize visual question answering (VQA) systems. VQA allows users to formulate free-form questions to extract generic information from remote sensing images.  It has been shown that the fusion of the input modalities (i.e., image and text) is crucial for the performance of VQA systems. Many current fusion approaches use modality-specific representations in their fusion modules instead of joint representation learning. However, to fully utilize both input modalities, the model has to learn the joint representation instead of combining the modality-specific representations. We present a multi-modal transformer architecture as a fusion module for VQA in remote sensing to overcome this issue. The presented multi-modal fusion transformer aims at learning a joint representation from both input modalities. Experimental results obtained on the two largest RS VQA datasets demonstrate the performance improvements of the joint vision and language transformers for the VQA task in RS.

\end{document}

%% file: introduction.tex
With advances in satellite technology, remote sensing (RS) image archives are rapidly growing, providing an unprecedented amount of data, which is a great source for information extraction in the framework of several different Earth observation applications. As a result, there has been an increased demand for systems that provide an intuitive interface to this wealth of information.
For this purpose, the development of accurate visual question answering (VQA) systems has recently become an important research topic in RS [\citenum{RSVQA}].
VQA defines a framework that allows retrieval of use-case-specific information [\citenum{VQA}].
By asking a free-form question to a selected image, the user can query various types of information without any need for remote sensing-related expertise.

Most of the existing VQA architectures consist of three main modules: i) the feature extraction module; ii) the fusion module; and iii) the classification module.
The feature extraction module extracts high-level features for both input modalities (i.e., image and question text).
After encoding the input modalities, the fusion module is required to discover a cross-interaction between both features.
Since the model needs to select the relevant part of the image concerning the question, combining the features in a meaningful way is crucial.
Finally, the output of the fusion module is passed to the classification module to generate the natural language answer.
In RS, relatively few VQA architectures are investigated [\citenum{RSVQA,cross-modal,MutualAtt,three_class_heads,mcb_mutan_rs,Chappuis_2022_CVPR}].
In [\citenum{RSVQA}], a convolutional neural network (CNN) is applied as an image encoder in the feature extraction module.
The natural language encoder is based on the skip-thoughts [\citenum{skip_thoughts}] architecture, while the fusion module is defined based on a simple, non-learnable point-wise multiplication of the feature vectors.
In [\citenum{MutualAtt}], image modality representation is provided by a VGG-16 network [\citenum{VGG_net}], which extracts two sets of image features: i) an image feature map extracted from the last convolutional layer; and ii) an image feature vector derived from the final fully connected layer.
The text modality representations are extracted by a gated-recurrent unit (GRU) [\citenum{GRU}].
The two modality-specific representations are merged by the fusion module that leverages the more complex mutual-attention component.
In [\citenum{mcb_mutan_rs}], the authors reveal the potential of an effective fusion module to build competitive RS VQA models.
In the computer vision (CV) community, it has been shown that to discover the underlying relation between the modalities, a model is required to learn a joint representation instead of applying a simple combination (e.g., concatenation, addition, or multiplication) of the modality-specific feature vectors [\citenum{VisualBERT}].
%This potential can already be seen in state-of-the-art CV-VQA architectures.
%These architectures indicate that to discover the underlying relation between the modalities, a model is required to learn the joint representation instead of applying a simple combination (e.g., concatenation, addition, or multiplication) of the feature vectors. \todo{review}
In [\citenum{cross-modal}], this knowledge is transferred to the RS VQA domain by introducing the first transformer-based architecture combined with an object detector pre-processor for the image features (called \emph{CrossModal}).
The object detector is trained on an RS object detection dataset (i.e., the xView dataset [\citenum{xView}]) to recognize target objects such as cars, buildings, and ships.
However, since the CrossModal architecture leverages objects defined in xView, the model is specialized for VQA tasks that contain those objects. In contrast, the largest RSVQA benchmark dataset (RSVQAxBEN [\citenum{RSVQAxBEN}]) includes questions regarding objects not included in xView (e.g., the dataset contains questions that require a distinction between coniferous and broad-leaved trees).
In [\citenum{Chappuis_2022_CVPR}], a model called  \emph{Prompt-RSVQA} is proposed that takes advantage of the language transformer DistilBERT [\citenum{distilBERT}].
To this end, the image feature extraction module uses a pre-trained multi-class classification network to predict image labels, then projects the image labels into a word embedding space. Finally, the image features encoded as words and the question text are merged in the transformer-based fusion module. Using word embeddings of the image labels as image features in Prompt-RSVQA comes with limitations: i) the model is tailored to the dataset that the classifier is trained on; ii) some questions are not covered in this formulation (e.g., counting, comparison) since the relation between the class labels are not presented in the predicted label set.
The aforementioned models provide promising results, guiding the motivation towards fusion modules that exploit transformer-based architectures. However, they are limited to the datasets that the classifier or the object detector is trained on, which makes the model unable to learn questions that are not covered in this formulation (e.g., counting, comparison, and objects that are not in the classifier/object detector training set).

To overcome these issues, in this paper, we present a multi-modal transformer-based architecture that leverages a different number of multi-modal transformer layers of the VisualBERT [\citenum{VisualBERT}] model as a fusion module.
% use: we dont need pre-defined objects and no need for labels -> see the Cross-Modal & Prompt-RSVQA description
% do not overemphasize the name just give a rough idea of the limitation  that our model is "solving"
Instead of applying an object detector, we utilize \emph{BoxExtractor}, a more general box extractor which does not overemphasize objects, within the image feature extraction module.
The resulting boxes are fed into a ResNet [\citenum{ResNet}] to generate image tokens as embeddings of the extracted boxes.
For the text modality, the BertTokenizer [\citenum{BERT}] is used to tokenize the questions.
Our fusion module takes modality-specific tokens as input and processes them with a user-defined number $l$ of VisualBERT (VB) layers.
The classification module consists of a \gls{mlp} that generates an output vector that represents the answer.
% The proposed architecture, called \emph{VBFusion}, is capable of handling various types of questions and is not limited to questions concerning pre-defined objects..
Experimental results obtained on large-scale RS VQA benchmark datasets [\citenum{RSVQAxBEN,RSVQA}] (which only include the RGB bands of Sentinel-2 multispectral images) demonstrate the success of the proposed architecture (called \emph{VBFusion}) compared to the standard RS VQA model. 
To analyze the importance of the other spectral bands for characterization of the complex information content of Sentinel-2 images in the framework of VQA, we add the other 10m and all 20m spectral bands to the RSVQAxBEN [\citenum{RSVQAxBEN}] dataset.
The results show that the inclusion of these spectral bands significantly improves the VQA performance, as the additional spectral information helps the model to take better advantage of the complex image data.
To the best of our knowledge, this is the first study to consider multispectral RS VQA that is not limited to the RGB bands.

The remaining sections are organized as follows; \cref{sec:method} will introduce the proposed architecture, including the feature extraction pipeline.
In \cref{sec:exp}, we will introduce the datasets for the experiments, the experimental setup, and in \cref{sec:experimental_results} analyze our results.
We will conclude our work in \cref{sec:con} and provide an outlook for future research directions.

%% file: method.tex
% Agreed to use l-layers inside of figures
\begin{figure}[tb!]
    \centering
    \includegraphics[width=.95\linewidth]{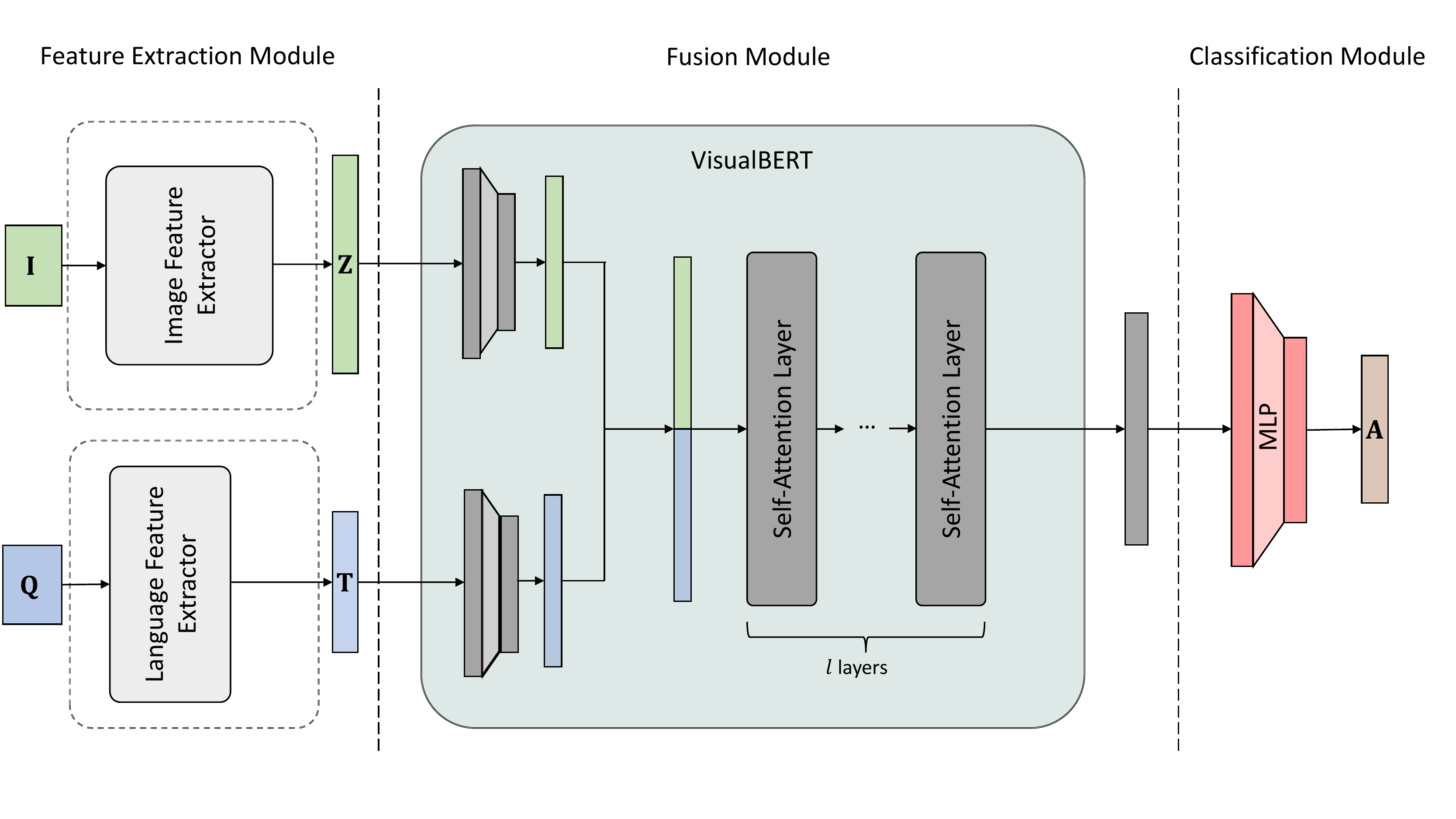}
    \caption{A general overview of VBFusion, our multi-modal transformer-based VQA architecture.
    Our architecture includes: i) a feature extraction module (\cref{feature_extraction_module}); ii) a fusion module (\cref{fusion_module}); and iii) a classification module (\cref{classification_module}).
    An RS image \textbf{I} and a question \textbf{Q} about this image are considered as input.
    Inside the VisualBERT model, the modality-specific features [\textbf{Z}, \textbf{T}] get projected to a hidden dimension and concatenated afterwards.
    The concatenated features are then fed to $l$ Self-Attention Layers, providing an output for further processing in the classification module.
    The output of the classification module is a vector representing the final answer \textbf{A} to the question \textbf{Q}.}
    \label{overview_model}
\end{figure}

%In this section, we present the details of our proposed multi-modal fusion transformer.
Visual question answering systems attempt to answer questions in natural language regarding an image input and can be formulated as a classification problem, where image and question are the input and the answer is considered as a label.
Given an image, question, answer triplet denoted as $(\mathbf{I, Q, A})$, an VQA system inputs $(\mathbf{I, Q})$ and predicts $\mathbf{A}$.
We assume that a training set $\mathcal{D} = \{\mathbf{I}_n, \mathbf{Q}_n^m, \mathbf{A}_n^m\}^{m=1..M_n}_{n=1..N}$ that consists of $N$ number of images is available, and for each $n \in \{1,..., N\}$ the image $\mathbf{I}_n$ is associated with $M_n$ pairs of questions $\{\mathbf{Q}_n^m\}_{m=1}^{M_n}$ and answers $\{\mathbf{A}_n^m\}_{m=1}^{M_n}$. 
Given the training set $\mathcal{D}$, the proposed multi-modal fusion transformer aims to learn a joint representation (from images and questions) and a classification head to obtain the answer.
To this end, the proposed architecture includes: i) a feature extraction module based on the BoxExtractor and the BertTokenizer; ii) a fusion module based on a user-defined number of multi-modal transformer layers of VisualBERT; and iii) a classification module consisting of an \gls{mlp} projection head.
\cref{overview_model} shows the general overview of our multi-modal transformer-based VQA architecture VBFusion.
Each module is explained in detail in the following.

% TOOD: Discuss if we even need subsectiosn
\subsection{Feature Extraction Module}
\label{feature_extraction_module}
The feature extraction module aims to extract relevant features for text and image modality independently from each other. To this end, the feature extraction module consists of two modality-specific encoder networks: \linebreak i) image modality encoder $f$; and ii) text modality encoder $g$.
In the case of the image modality, we first utilize a simplified box extractor followed by an image encoder network. Then the output of the image encoder network is transformed with an \gls{mlp} layer into a suitable shape for the fusion module. It is worth noting that, to extract the image features, in CV it is common to guide the focus of a model to the relevant areas of the image [\citenum{VisualBERT, BottomUp}]. For this reason, the state-of-the-art VQA models in CV leverage an object detector to focus on important regions (i.e., objects). For example, in VisualBERT [\citenum{VisualBERT}] a Faster R-CNN [\citenum{FasterRCNN}] backbone is used to extract the relevant regions of the image. To extract regions of interest and detect objects present in RS images, semantic segmentation models are often applied. Such models require the availability of reliable pixel-based ground reference samples to be used in the training phase. 
The collection of a sufficient number of reliable labeled samples is time-consuming, complex, and costly in operational scenarios and can significantly affect the final accuracy of object detection. To overcome this issue, we propose \emph{BoxExtractor} which generates rectangular boxes, selecting a region of interest without requiring any labeled training samples.

\begin{figure}[tb!]
    \centering
    \includegraphics[width=.9\linewidth, trim=0 100pt 0 0, clip]{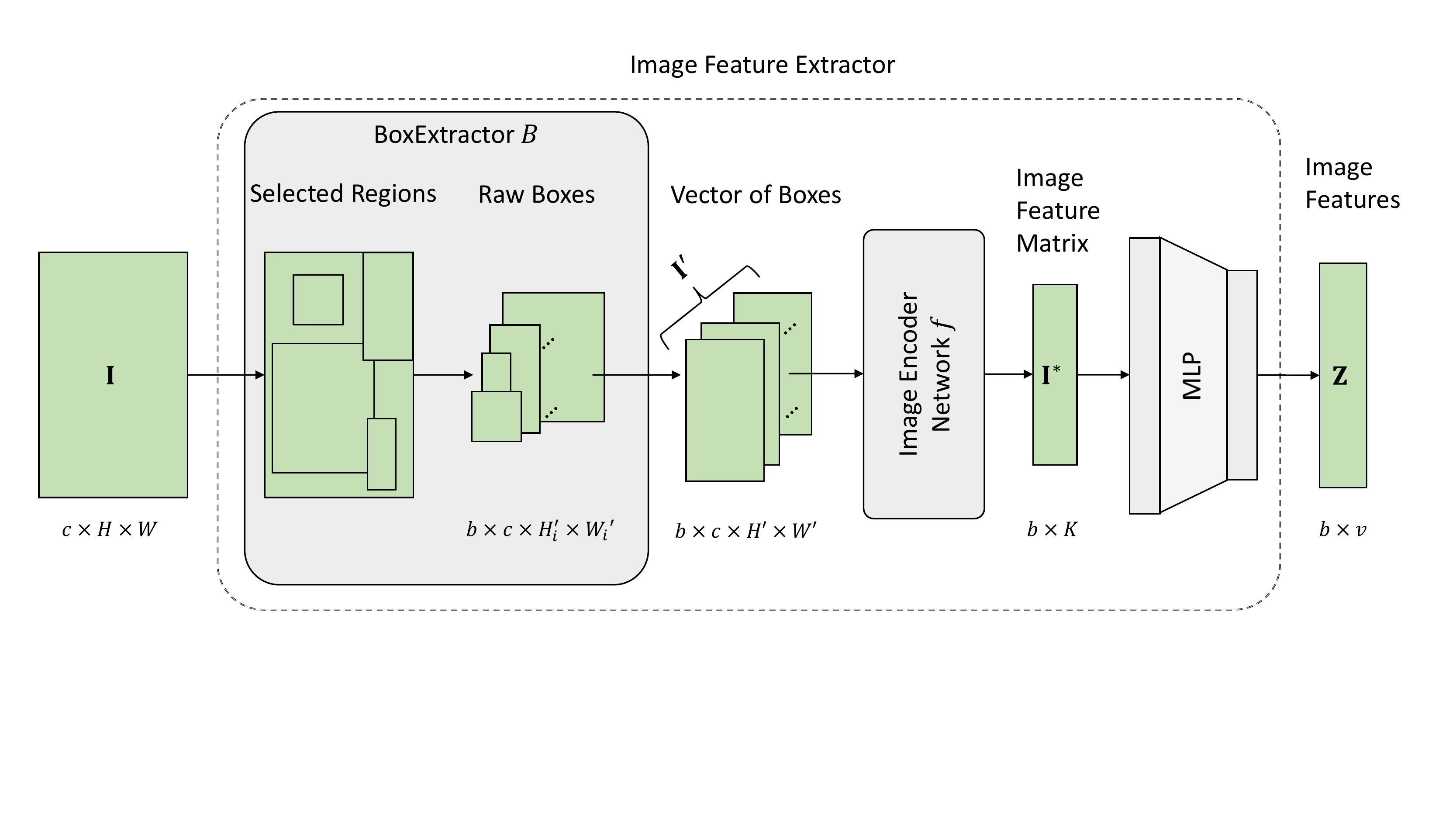}
    \caption{The image feature extraction module takes an image \textbf{I} as input and outputs a feature map \textbf{Z}. First, the BoxExtractor $B$ selects regions of the input image \textbf{I} and creates $b$ raw boxes, which has the sizes $c \times H_i^\prime \times W_i^\prime$ with $i=1,...,b$. 
    These raw boxes are interpolated to the size $c \times H^\prime \times W^\prime$ resulting in the vector of boxes $\textbf{I}^\prime=B(\textbf{I})$. To further process $\textbf{I}^\prime$, the image encoder network $f$ is leveraged. 
    The output of $f$, $\textbf{I}^\star$ (size $b \times K$) is then projected to image features \textbf{Z} (size $b \times v$) through an \gls{mlp} layer.}
    \label{visual_feature_extraction}
\end{figure}

The image feature extraction pipeline is illustrated in \cref{visual_feature_extraction}.
To obtain the image features, we first create spatially interpolated rectangular regions of the image $\mathbf{I}_n$ with our BoxExtractor. For this purpose, let $c \in \mathbb{N}$ be the number of spectral bands, $H, W \in \mathbb{N}$ the height and width of the image, and $H', W' \in \mathbb{N}$ the height and width of the interpolated boxes.
Let $b \in \mathbb{N}$ be the number of boxes that BoxExtractor creates.
Then the function $B: \mathbb{R}^{ c \times H \times W} \to \mathbb{R}^{b \times c \times H' \times W'}$ defines the BoxExtractor that maps the image $\mathbf{I}_n$ to its boxes $B(\mathbf{I}_n) = \mathbf{I}_n^\prime$ in a non-deterministic way.
The non-determinism is observed because we first choose random start- and end-points for the height and the width of the boxes, $s_H^i, e_H^i \in \{0, 1, ..., H\}$, $s_H^i < e_H^i$ and $s_W^i, e_W^i \in \{0, 1, ..., W\}$, $s_W^i < e_W^i$, for $i \in \{1, ..., b\}$. In the second step, a vector of boxes to each pair of start- and end-points is created.
We control the distance between the start- and endpoints (e.g., $e_H^i - s_H^i$) to guarantee a minimum amount of information in each box.
To this end, we set:
\begin{equation}
    \text{min}_H \leq e_H^i - s_H^i 
\end{equation}
and
\begin{equation}
    \text{min}_W \leq e_W^i - s_W^i,
\end{equation}
for $i = 1, ..., b$, where $\text{min}_H, \text{min}_W$ are hyperparameters.
The resulting vector of boxes $\textbf{I}_n^\prime$ is obtained by equally resizing the boxes to $H'$ and $W'$ with bicubic interpolation.  Then, we stack the image boxes and pass them to the image encoder network $f$.
Finally, we transform the output $\textbf{I}^*=f(\textbf{I}_n^\prime)$ into a suitable shape for the fusion module.
To this end, we reshape the features and keep the number of boxes as the first dimension, whereas the second dimension is given by flattening the remaining directions, resulting in a matrix.
The matrix is fed into an \gls{mlp}, to project its size to $b \times v$, where $v \in \mathbb{N}$ is the so-called image embedding dimension of the VisualBERT [\citenum{VisualBERT}] model (i.e., the required input dimension for the image modality).
Finally, the output image embedding vectors $\mathbf{Z}_n=\mathrm{MLP}(\textbf{I}^*)$ are given to the fusion module.
%Note that, except for the last linear layer, the image feature extraction module is frozen and may be applied as a pre-processing step.
% In the actual code, the linear layer is shifted to the fusion model. % <- irrelevant for the paper
% In the case of 10 bands, $b=10$, we have to resize the spectral bands before the box extraction to have all bands in the same resolution. % <- Implementation detail that is 'well-known' when working with multi-spectral data; doesn't fit the narrative here

To extract the text features from a question $\mathbf{Q}_n^m$, we translate the question into tokens. These tokens are a mapping of the words to a numerical value.
To achieve this and to be aligned with the original VisualBERT model, we utilize a frozen BertTokenizer [\citenum{BERT}]\, as text modality encoder network $g$, where $\mathbf{T}_n^m=g(\mathbf{Q}_n^m)$ is the output text embedding vector. Both $\mathbf{Z}_n$ and $\mathbf{T}_n^m$ is jointly processed to extract their implicit knowledge in the next step.

\subsection{Fusion Module}
\label{fusion_module}
Our transformer-based fusion module jointly learns the alignment between the image and text modality.
To learn the joint alignment, we leverage $l$ VisualBERT layers.
These layers are an extension of the language transformer BERT with the image modality.
VisualBERT first projects the image features $\mathbf{Z}_n$ into a suitable shape to concatenate the projected image features with the language modality.
The concatenated image and text features $[\mathbf{Z}_n, \mathbf{T}_n^m]$ are then fed to the BERT transformer layers.
The output is given by pooling the last hidden state.
The resulting vector is further processed in the classification module.
To summarize the forward procedure, VisualBERT fuses the image and language modalities and feeds them into the BERT architecture, representing a natural extension of BERT to multiple modalities.
Additionally, VisualBERT shows competitive results on VQA benchmark datasets and is thus well suited as a starting point for multi-modal fusion transformers in RS VQA.
%Unlike the previous RS VQA models that make use of simple combination of both modalities
Unlike the previous RS VQA models that simply combine both modalities in a non-learnable fashion, our model learns a joint representation.
% Our proposed RS VQA architecture contraries the previous RS VQA models, which combine both modalities in a non-learnable way.
Note that our approach differs from [\citenum{Chappuis_2022_CVPR}] because they apply a transformer only to the language modality. In contrast to the model proposed in [\citenum{cross-modal}], we apply exclusively multi-modal self-attention instead of a mixture of single- and multi-modal self-attention. 

\subsection{Classification Module}
\label{classification_module}

The task of the classification module is to create an output vector representing the prediction for the specific answers.
Following [\citenum{RSVQA}], we utilize an \gls{mlp} as a classification module.
The \gls{mlp} consists of three layers, where the last layer has the dimension of the considered answer set.
The final answer is then obtained by selecting the largest activation of the output vector.
% Remark that the answers are declining sorted in a vector by their occurrence frequency, and the position of the answers in this sorted vector is the corresponding entries of the classification vector. % <= Also not relevant for this section

%% file: experiments.tex
We conducted experiments on the RSVQA-LR [\citenum{RSVQA}] and the RSVQAxBEN [\citenum{RSVQAxBEN}] datasets.
The RSVQA-LR dataset was constructed using 7 Sentinel-2 tiles acquired over the Netherlands, from which only the RGB bands were used.
The tiles are divided into 772 patches of size 256x256 pixels.
To obtain the information needed to create the image-question-answer triplets, knowledge from the publicly available OpenStreetMap database was leveraged.
With this knowledge, the dataset was constructed by taking an image patch and randomly generating question-answer pairs from a given template.
This procedure resulted in 77,232 image-question-answer triplets.
As in [\citenum{RSVQA}], we used the same tile-based train/validation/test split.
The training split consists of five tiles, whereas the validation and test split consist of one tile each. 
For further information on the RSVQA-LR dataset (e.g., the definition of the question classes), the reader is referred to [\citenum{RSVQA}].

The RSVQAxBEN [\citenum{RSVQAxBEN}] dataset embodies the largest freely available RS VQA benchmark dataset.
It is based on the BigEarthNet (BEN) [\citenum{BEN}] archive and contains 590,326 Sentinel-2 L2A image patches with 12 spectral bands of varying resolution (10m, 20m, and 60m).
The content of each patch is described by multiple class labels from the CORINE Land Cover (CLC) 2018 database.
The RSVQAxBEN dataset was constructed as a large-scale benchmark for RS VQA, using only the RGB bands of the BEN patches (which have a spatial resolution of 10m). A stochastic algorithm that utilizes the CLC labels produced the image-question-answer triplets.
% A stochastic algorithm, that took advantage of the CLC labels, produced the image-question-answer triplets.
The procedure described in [\citenum{RSVQAxBEN}] resulted in 14,758,150 image-question-answer triplets,
which include 26,875 unique answers.
To limit the number of possible answers, the model's output in [\citenum{RSVQAxBEN}] was restricted to the 1,000 most frequent answers, covering \SI{98.1}{\percent} of the answer set.
For the sake of comparability, we also apply this restriction.
We use the same train/validation/test split based on the tiles' spatial location as in [\citenum{RSVQAxBEN}].
For more statistical insights, the reader is referred to [\citenum{RSVQAxBEN}]\,.
We provide experimental results on the original RSVQAxBEN dataset. 
Furthermore, we extended the original three-band RSVQAxBEN dataset to a ten-band version by including all spectral bands with 10m and 20m spatial resolution in addition to the RGB bands.
In our experiments, we resampled the 20m bands to the resolution of the 10m bands by using a cubic interpolation method.

We performed experiments with different layer configurations for our proposed VBFusion architecture given by the pre-trained VisualBERT [\citenum{VisualBERT}] model. The value of $l$ (which defines the number of layers selected from the original VisualBERT model) is varied as $l$ $\in\{4, 6, 8, 10, 12\}$. 
Note that when $l=12$, it is identical to the original VisualBERT model. 
For the image encoder network $f$, we utilized a pre-trained ResNet152 [\citenum{ResNet}] network architecture with frozen weights. 
The image encoder network was pre-trained on ImageNet [\citenum{ImageNet}]\, for RSVQA-LR and three-band RSVQAxBEN and on BEN for the ten-band variant. 
We extracted ten boxes for the RSVQA-LR and three-band RSVQAxBEN datasets, and for the ten-band variant five boxes.
The learning rate was set to $10^{-6}$, while the maximum number of training epochs was set to 300 for RSVQA-LR and 20 for RSVQAxBEN. We chose a batch size of 1024 and 2048 for RSVQA-LR and RSVQAxBEN, respectively.
We performed our experiments on a cluster of 8 NVIDIA Tesla A100 GPUs.
We used the implementation from the Huggingface library [\citenum{wolf-etal-2020-transformers}] for the VisualBERT layers.
Our architecture was compared with the SkipRes [\citenum{RSVQAxBEN}] architecture, which is based on Skip-Thoughts and ResNet152. To analyze the results on the RSVQA-LR dataset we consider the same metrics as given in [\citenum{RSVQA}]:  the accuracy of counting questions (called \enquote{Count}), of presence questions (called \enquote{Presence}), of comparison questions (called \enquote{Comparison}) and of rural/urban questions (called \enquote{Rural/Urban}). Furthermore, we provide the average accuracy (\enquote{AA}) over the question type accuracies and the overall accuracy (\enquote{OA}).
The metrics used to analyze the results on the RSVQAxBEN dataset are: the accuracy of yes/no questions (called \enquote{Yes/No}) and the accuracy of the land use and land cover questions (called \enquote{LULC}), as well as \enquote{OA} and \enquote{AA}.

\begin{table}[tb!]
    \centering
    \caption{Accuracies obtained by using SkipRes and VBFusion with different number $l$ of layers (RSVQA-LR dataset).}
    % \begin{tabular}{l c K K K K K K}
    \begin{tabular}{l c   c c c c c c}
    \toprule
    \multirow[b]{2}*{Architecture} & \multirow[b]{2}*{$l$} & \multicolumn{4}{c}{Question Type} & {\multirow[b]{2}*{AA}} & {\multirow[b]{2}*{OA}} \\
    \cmidrule[0.025em](lr){3-6}
    {} & {} & {Count} & {Presence} & {Comparison} & {Rural/Urban} & {} & {} \\
    \midrule
    SkipRes [\citenum{RSVQAxBEN}] & {--} & 67.01 & 87.46 & 81.50 & \textbf{90.00} & 81.49 & 79.08 \\ \cmidrule[0.025em](lr){2-8}
    \multirow[c]{5}*{VBFusion} & 4 & 68.17 & 88.02 & 88.51 & 87.00 & 82.92 & 82.36 \\
     & 6 & 68.17 & 88.87 & \textbf{88.83} & 84.00 & 82.47 & 82.71 \\
     & 8 & \textbf{69.36} & 89.00 & 83.46 & 88.00 & 82.45 & 80.99 \\
     & 10 & 67.73 & \textbf{89.48} & 86.68 & 88.00 & \textbf{82.97} & 81.94 \\
     & 12 & 68.14 & 89.27 & 88.71 & 85.00 & 82.78 & \textbf{82.78} \\
    % CrossModal & \textbf{70.06} & \textbf{91.85} & \textbf{93.07} & 50.00 & 76.21 & \textbf{85.50} \\
    \bottomrule
    \end{tabular}
    \label{RSVQA_LR_acc}
\end{table}

\section{Experimental Results}
\label{sec:experimental_results}

\cref{RSVQA_LR_acc} shows the results obtained  on the RSVQA-LR dataset. By analyzing the table, one can see that our proposed VBFusion architecture outperforms the SkipRes model in all metrics except \enquote{Rural/Urban}, independently of the number of layers.
The most significant difference occurs in the \enquote{Comparison} accuracy, which represents the most challenging question type. 
Here, a model is required to count the number of occurrences of a selected object with a specific positional relationship to a reference object.
For this type of question, our 6-layer VBFusion architecture significantly outperforms SkipRes by more than \SI{7}{\percent}.
For the \enquote{Count} questions the differences are smaller, e.g., VBFusion with 10 layers outperforms the SkipRes model by only \SI{0.7}{\percent}.
In the summarizing metrics \enquote{AA}, and \enquote{OA}, all VBFusion models, independently of the number of layers, significantly outperform SkipRes.
% VBFusion with 10 layers beats SkipRes by more than \SI{3}{\percent} in \enquote{AA}.
% For \enquote{OA} we obtain just a slight improvement for our models, except for VBFusion with 8 layers, which underperforms the SkipRes.
Furthermore, no significant improvement is observed when increasing the number $l$ of VisualBert layers.
% Compared to the relatively large complexity of our models, the slight improvement is aligned with the common knowledge that big transformers\cite{VIT} struggle on small datasets.
Although complexity increases, when using more layers, performance changes only slightly.
This observation aligns with the findings of [\citenum{VIT}] that large transformers struggle on small datasets.
To benefit from the larger transformer configurations, more samples are probably required.
In the case of the \enquote{Rural/Urban} questions, the underperformance can be explained with the same argument; there are too few questions of this type.
This is because the \enquote{Rural/Urban} question type makes up the smallest proportion of the dataset (\SI{1}{\percent}).
By analyzing the results, one can conclude that the proposed VBFusion architecture is able to improve the performance compared to the SkipRes architecture. It is worth emphasizing that the datasets used in the experiments are benchmarks, whereas in many real applications the VQA is expected to be applied to much larger archives. Due to the nature of the large transformer-based architecture, we expect that the gain achieved by the VBFusion architecture will be increased for large-scale RS VQA problems and also for the cases of when much more complex question types are present.

% The Rural/Urban question-type is the smallest question type that only takes up \SI{1}{\percent} of the dataset.
%When comparing our results to CrossModal, our model underperforms in OA because most questions are concerned with predictable objects.
%However, our model has a higher AA score, which underlines the general characteristics of our proposed architecture.

\begin{table}[tb!]
    \centering
    \caption{Accuracies obtained by using SkipRes and VBFusion with different number $l$ of layers (RSVQAxBEN dataset).}
    \label{RSVQAxBEN_acc}
    % \begin{tabular}{c c c K K K K}
    \begin{tabular}{c c c c c c c}
        \toprule
        \multirow[b]{2}*{Architecture} & \multirow[b]{2}*{$l$}
            & \multirow[b]{2}*{Number of Bands} & \multicolumn{2}{c}{Question Type} & {\multirow[b]{2}*{AA}} & {\multirow[b]{2}*{OA}} \\
        \cmidrule[0.025em](lr){4-5}
        {} & {} & {} & {LULC} & {Yes/No} & {} & {} \\
        \midrule
        SkipRes [\citenum{RSVQAxBEN}] & {--} & 3 & 20.68 & 80.02 & 50.35 & 69.92  \\
        \cmidrule[0.025em](lr){2-7}
        \multirow{10}*{VBFusion} & \multirow{2}*{4} & 3 &  20.40 & 83.86& 52.13 & 73.06 \\
        {} & {} & 10 & 25.72 & 85.41 & 55.56 &  75.26 \\
        \cmidrule[0.025em](lr){3-7}
        {} & \multirow{2}*{6} & 3 & 21.66 & 84.58 & 53.12 &  73.88 \\
        {} & {} & 10 & 25.88 & 85.48 & 55.68 & 75.34  \\
        \cmidrule[0.025em](lr){3-7}
        {} & \multirow{2}*{8} & 3 & 20.07 & 84.37 & 52.22 &  73.43 \\
        {} & {} & 10 & 25.04 & \textbf{86.56} & \textbf{55.80} & \textbf{76.10} \\
        \cmidrule[0.025em](lr){3-7}
        {} & \multirow{2}*{10}& 3 & 20.82 & 85.13 & 52.97 & 74.19  \\
        {} & {} & 10 &  25.19 & 85.95 & 55.57 & 75.61  \\
        \cmidrule[0.025em](lr){3-7}
        {} & \multirow{2}*{12}& 3 & 24.33  & 85.47 & 54.90 & 75.07 \\
        {} & {} & 10 & \textbf{26.26} & 85.34 & \textbf{55.80} & 75.29 \\
        \bottomrule
    \end{tabular}
\end{table}

% Text from recommended paper:
% https://arxiv.org/pdf/2006.11529.pdf
%  Tables IV and V report the classification
% accuracies and computational times for the NWPU-RESISC45
% and AID image archives, respectively.
% By analyzing the tables one can observe
% that the [...]
% In addition, we can also notice that the proposed
% approach attains [...]
% On the contrary, if we perform classification [...]
% By analyzing the AlexNet model results for NWPURESISC45 archive (Table IV), we can notice that the classification accuracy obtained by
% , the required classification time is of more than an order of magnitude smaller at the cost of almost 5 lower classification accuracy
% When we compare the performance of the proposed approach with the standard-CNN, although [...], there is a signifcant gain in terms of ...
% By analyzing the results, one can conclude that the proposed approach minimizes YYY considerably when compared to XXX.
% In addition, by using ZZZ the performance is also improved.
% However, this is achieved at the cost of TTT.

\Cref{RSVQAxBEN_acc} reports the accuracies for the three- and ten-band RSVQAxBEN dataset variants.
% The results for the RSVQAxBEN dataset are presented in \cref{RSVQAxBEN_acc}. 
By analyzing the tables one can observe that the proposed architecture VBFusion outperforms the SkipRes architecture, irrespective of the layer configuration, in both summarizing metrics \enquote{AA} and \enquote{OA}.
Even the smallest configuration with 4 layers trained on three bands improves the \enquote{AA} by almost \SI{2}{\percent} and the \enquote{OA} by more than \SI{3}{\percent}.
When we compare the performance of the proposed architecture trained on the three-band variant, most of the improvements are attributed to the \enquote{Yes/No} related questions.
We can also see that in the 3-band scenario, deeper and larger VBFusion models lead to higher \enquote{OA} scores.
% \todo{maybe better reason about this, but would require more discussion for the 10-band version too}
% Due to the large RSVQAxBEN dataset, the larger transformer models are able to better align the features
This pattern is in contrast to the generally similar performance in the \enquote{LULC} question category among the layer configurations up to $l=10$ trained on three bands.
Interestingly, a performance jump can be observed for the \enquote{LULC} questions with the largest layer configuration ($l=12$).
A possible reason is that with 12 layers, all pre-trained VisualBERT layers are utilized and none are rejected.
Therefore, the initial weights can produce more meaningful high-level embeddings than the pruned models.
This seems to be especially important for the complex \enquote{LULC} questions, where the model is required to deeply understand the contents of intricate multispectral images and connect them to the associated question.
The performance improvement exists for the three- and ten-band variants, although it is more dominant in the three-band configuration.
With all 12 layers, the \enquote{LULC} accuracy increases by \SI{3.5}{\percent}, compared to the 10 layer configuration, for the three-band training leading to a high \enquote{OA} of \SI{75.07}{\percent}.
The effect is not as significant for the 10-band configuration.
The relatively low impact on the 10-band configuration can be due to a discrepancy between the high-level RGB image feature representation and the ten-band feature representation from the initial weights of the pre-trained VisualBERT model, since VisualBERT was pre-trained on RGB images.
However, analyzing the results for all 10-band configurations, one can observe that these models improve the performance in all metrics compared to their three-band counterparts, except for the 12-layer configuration in the \enquote{Yes/No} category, where the accuracy is slightly lower by less than \SI{0.2}{\percent}.
Most of the observed performance increase is due to the notably higher accuracies in the \enquote{LULC} category.
For example, the smallest configuration trained on 10 bands with $l=4$ is more than \SI{5}{\percent} better than the same architecture trained on the 3 band variant in the \enquote{LULC} category.
An improved \enquote{LULC} performance is expected, since some LULC classes, such as different types of vegetation, greatly benefit from additional spectral information.
The smallest model trained on ten bands even outperforms the largest and best-performing configuration trained on three bands by \SI{0.6}{\percent} in \enquote{AA} and almost by \SI{0.2}{\percent} in \enquote{OA}.
% and significantly outperforms the SkipRes architecture by \SI{5.2}{\percent} and \SI{5.3}{\percent}, respectively.
When training with ten bands, it can be observed that larger configurations of our proposed VBFusion architecture do not necessarily lead to higher accuracies.
The best performing ten-band trained configuration utilizes 8 layers and reaches an \enquote{OA} of \SI{76.10}{\percent}, while the full 12 layer configuration has the second lowest \enquote{OA} with \SI{75.29}{\percent}.
By analyzing the results, one can conclude that the proposed VBFusion architecture discovers the underlying relationship between both the image and question modality better than models that utilize a simple feature combination as the fusion module.
Furthermore, the results show the importance of utilizing additional spectral bands, when available, to better model the contents of intricate multispectral imagery for VQA systems.

%% file: conclusion.tex
In this paper, we have presented a novel architecture in RS VQA that applies a transformer model, which exclusively relies on multi-modal transformer layers to learn the image and text representations jointly.
Our proposed architecture VBFusion includes: i) a feature extraction module based on the BoxExtractor, ResNet152, and the BertTokenizer; ii) a fusion module based on a user-defined number of multi-modal transformer layers of VisualBERT; and iii) a classification module consisting of an \gls{mlp} projection head.
It is worth noting that our architecture is not limited to specific questions, e.g., questions concerning pre-defined objects.
To show the effectiveness of the proposed architecture, we have evaluated the model on the RSVQA-LR dataset, the RSVQAxBEN dataset (which only includes the RGB bands of the Sentinel-2 image patches), and an extended RSVQAxBEN dataset (which includes all the spectral bands of Sentinel-2 images with 10m and 20m spatial resolution).
From the experimental results obtained on the RSVQAxBEN dataset variants, we observe that: i) our architecture leads to significant performance improvements compared to an architecture that simply combines the modality-specific representations in the fusion module as our architecture better discovers the underlying relationship between the modalities; 
% 2) increasing the size and depth of the transformer leads to further performance gains if the dataset is large enough; \todo{I would remove 2} 
and ii) exploitation of additional available spectral bands leads to better modeling of the complex spatial and spectral content of RS images in the context of VQA.
Our architecture results in a performance increase for the comparably small RSVQA-LR dataset compared to the SkipRes architecture, which utilizes a simple feature combination as a fusion module. Compared to the relatively large complexity of our models, the slight improvement aligns with the literature findings that large transformers struggle on small datasets. In addition, larger layer configurations do not provide a significant improvement in the RSVQA-LR dataset. These results show that our multi-modal transformer-based fusion module requires larger training sets to realize its full
potential.
% Our results show that our proposed multi-modal transformer-based fusion module is able to improve the performance of VQA models with
%However, regarding the big RS image archives there are large-scale datasets available.
% However, with the ever-increasing remote sensing archives, the sizes of future datasets are expected to increase.
% Requires a lot data <= Ok because RS a lot of data

% Kai: check again later

We would like to note that although transformer-based models have a potential to provide high VQA performance in RS, they are associated to a high number of model parameters and high training complexity. Thus, as a future work, we plan to investigate efficient transformers and conduct a comparative study in terms of complexity and performance. Furthermore, we plan to extend our feature extraction module to optimize the BoxExtractor or exchange it with a more sophisticated general box extractor.